\documentclass[journal]{IEEEtran}
\usepackage{bm}
\usepackage{amsmath}
\usepackage{amssymb}
\usepackage{amsthm}
\usepackage{mathrsfs}
\usepackage{enumerate}
\usepackage{multirow}
\usepackage{color}
\usepackage{subfigure}
\usepackage[square, comma, sort&compress, numbers]{natbib}
\usepackage[pagebackref=false,breaklinks=true,letterpaper=true,colorlinks,bookmarks=false]{hyperref}
\usepackage{times}
\usepackage{breakurl}
\usepackage{array}
\usepackage{verbatim}
\usepackage{algorithm}
\usepackage{algpseudocode}
  \usepackage[pdftex]{graphicx}
  \usepackage{graphics}
  \usepackage{graphicx}
  \usepackage{epsfig}
  \usepackage{epstopdf}
\ifCLASSINFOpdf
\else
\fi

\makeatletter
\def\hlinewd#1{%
  \noalign{\ifnum0=`}\fi\hrule \@height #1 \futurelet
   \reserved@a\@xhline}
\makeatother

\begin{document}
%
\title{Vehicle Re-identification Using Quadruple Directional Deep Learning Features}

\author{Jianqing Zhu, Huanqiang Zeng, Jingchang Huang, Shengcai Liao, Zhen Lei, Canhui Cai and LiXin Zheng
\thanks{This work was supported in part by the National Natural Science Foundation of China under the Grants 61602191, 61401167, 61605048, 61473291 and 61372107, in part by the Natural Science Foundation of Fujian Province under the Grants 2018J01090 and 2016J01308, in part by High-level Talent Innovation Program of Quanzhou City under the Grant 2017G027, in part by the Promotion Program for Young and Middle aged Teacher in Science and Technology Research of Huaqiao University under the Grants ZQN-PY418 and ZQN-YX403, in part by the Scientific and Technology Founds of Xiamen under the Grant 3502Z20173045, and in part by the Scientific Research Funds of Huaqiao University under the Grant 16BS108, 14BS201 and 14BS204. \emph{(Corresponding author: Huanqiang Zeng)}.
}
\thanks{Jianqing Zhu, Canhui Cai and LiXin Zheng are with the College of Engineering, Huaqiao University, Quanzhou, 362021, China and Fujian Provincial Academic Engineering Research Centre in Industrial Intellectual Techniques and Systems (e-mail: \{jqzhu, chcai, zlx\}@hqu.edu.cn).}
\thanks{Huanqiang Zeng is with the College of Information Science and Engineering, Huaqiao University, Xiamen, 361021, China (e-mail: zeng0043@hqu.edu.cn).}
\thanks{Jingchang Huang is with IBM-Research China, Building 10/6F, 399 Ke-Yuan Road, Zhangjiang Hi-Tech Park, Pudong New District, Shanghai 201203, China (e-mail: hjingc@cn.ibm.com).}
\thanks{Shengcai Liao and Zhen Lei are with the Center for Biometrics and Security Research and National Laboratory of Pattern Recognition, Institute of Automation, Chinese Academy of Sciences, Beijing 100190, China (e-mail: \{scliao, zlei\}@cbsr.ia.ac.cn).}
}



\maketitle

\begin{abstract}
 In order to resist the adverse effect of viewpoint variations
 for improving vehicle re-identification performance, we design quadruple directional deep learning networks to
 extract quadruple directional deep learning features (QD-DLF) of vehicle images. The quadruple directional
 deep learning networks are with similar overall architecture, including the same basic deep learning architecture but different
 directional feature pooling layers. Specifically, the same basic deep learning architecture is a shortly and densely connected
 convolutional neural network to extract basic feature maps of an input square vehicle image in the first
 stage. Then, the quadruple directional deep learning networks utilize different directional pooling layers,
 i.e., horizontal average pooling (HAP) layer, vertical average pooling (VAP) layer, diagonal
 average pooling (DAP) layer and anti-diagonal average pooling (AAP) layer, to compress the basic
 feature maps into horizontal, vertical, diagonal and anti-diagonal directional feature maps, respectively.
 Finally, these directional feature maps are spatially normalized and concatenated together as a quadruple
 directional deep learning feature for vehicle re-identification. Extensive experiments on both VeRi and
 VehicleID databases show that the proposed QD-DLF approach outperforms multiple state-of-the-art
 vehicle re-identification methods.

\end{abstract}


%
\IEEEpeerreviewmaketitle

\section{Introduction}
\begin{figure}[t]
    \centering
    \includegraphics[width=.9\linewidth]{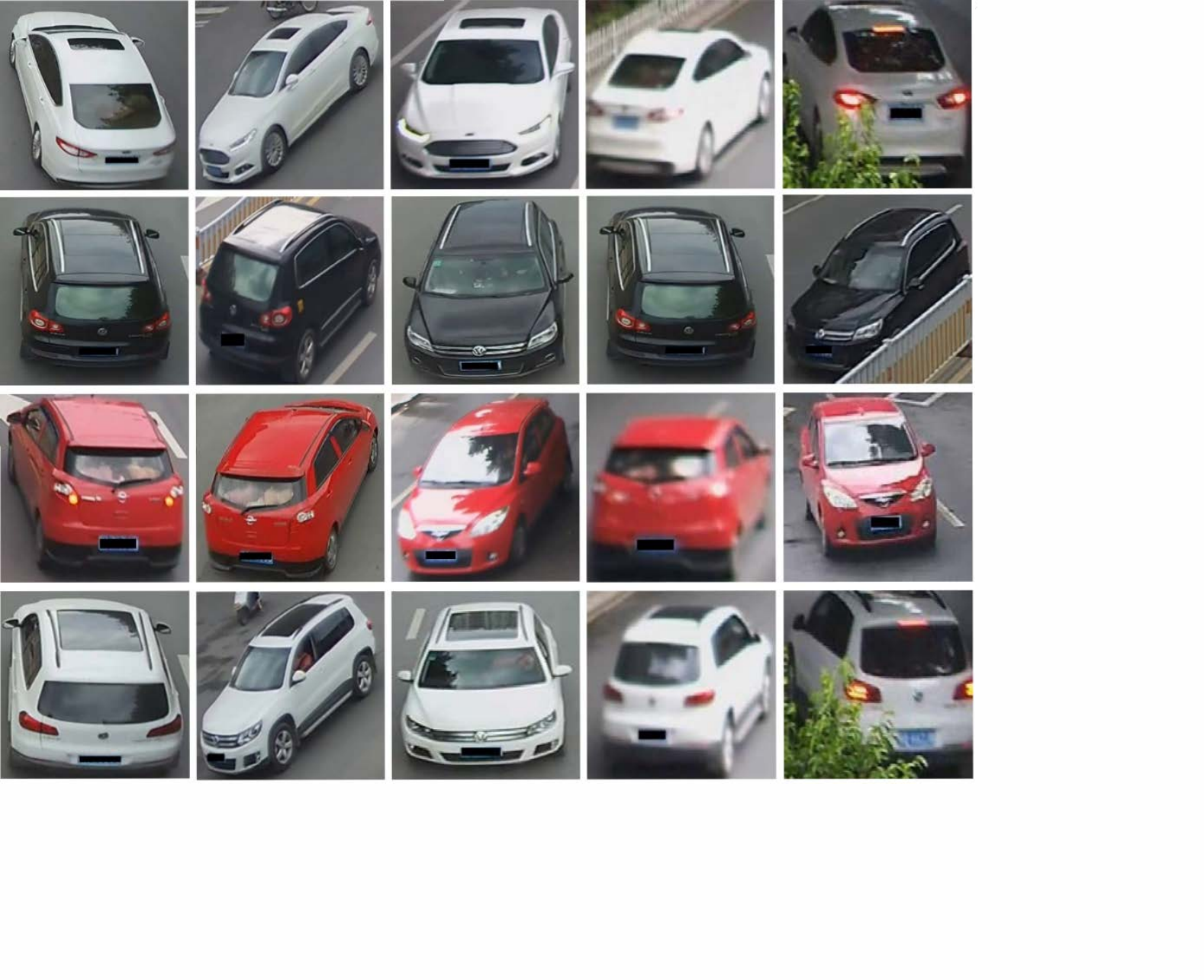}
    \caption{Classical vehicle samples from the VeRi \cite{veri2017} database. Each row denotes the same vehicle captured by cameras from different viewpoints.}
    \vspace{-.4cm}
    \label{fig:example}
\end{figure}

 Vehicle re-identification with the goal of matching the same vehicle image captured by different cameras plays an important role in video surveillance for public security, since vehicle has been an indispensable part of human daily life \cite{veri2017}.
 In practical scenarios, vehicle re-identification is a very challenging computer vision problem, due to the fact that vehicle images usually contain a lot of adverse factors, such as viewpoint variation, illumination change, blur, occlusion and low resolution, as shown in Fig. \ref{fig:example}. Therefore, how to design an effective vehicle re-identification method has attracted more and more attentions.

To address the problem of vehicle re-identification, two large benchmark databases, VeRi \cite{veri2016, veri2017} and VehicleID \cite{drdl}, are released by the Institute of Digital Media, Peking University. Based on these two databases, multiple vehicle re-identification methods were developed and will be highlighted in the following Section \ref{related}.
Note that for vehicle re-identification, viewpoint variation is most crucial challenge and often encountered factor among those above-mentioned adverse ones, because vehicle images are usually captured under different camera viewpoints. Hence, this paper focuses on designing a method to resist adverse viewpoint variations so as to improve the vehicle re-identification performance. For that, we propose quadruple directional deep learning features (QD-DLF) to comprehensively describe vehicle images for improving the vehicle re-identification performance.
The novelty and contribution of the proposed QD-DLF is that we make the first attempt to fuse quadruple directional deep features learned by using quadruple directional average pooling layers to improve the robustness of viewpoint variations.
Experimental results have shown that the proposed approach is able to significantly improve the vehicle re-identification accuracy.

The rest of this paper is organized as follows. Section \ref{related} introduces the related work. Section \ref{sec:method} describes the proposed quadruple directional deep learning features for vehicle re-identification. Section \ref{sec:exp} presents the experimental results to validate the superiority of the proposed method. Section \ref{sec:con} concludes this paper.

\section{Related Works}\label{related}

In this section, we briefly review the existing vehicle re-identification works. Since feature representation and similarity metric are two key roles in vehicle re-identification, the existing vehicle re-identification works mainly focus on two aspects: (1) feature representation for vehicle re-identification, and (2) similarity metric for vehicle re-identification.

\subsection{Feature Representation for Vehicle Re-identification}
Feature representation methods for vehicle re-identification can be mainly divided into two classes: hand crafted feature representations and deep
 learning features representations. For hand crafted feature representations, LOMO \cite{lomo} and BOW-CN \cite{market} features originally used in person re-identification are directly applied to vehicle re-identification. For deep learning feature representations, some well-known deep feature learning networks, such as AlexNet \cite{alexnet}, VGGNet \cite{vgg} and GoogLeNet \cite{googlenet}, are used as feature extractors for vehicle re-identification. For example, FACT \cite{veri2016} uses AlexNet \cite{alexnet} to extract features of vehicles. NuFACT \cite{veri2017} takes GoogLeNet \cite{googlenet} as a feature extractor. DRDL \cite{drdl} utilizes VGGNet \cite{vgg} to extract features of vehicles. It can be further found that the deep learning features obviously outperforms hand crafted features on VeRi and VehicleID databases as reported in~\cite{veri2016, veri2017, drdl}.

To effectively learn discriminative feature representations of vehicle images, multiple types of loss functions are applied to train deep learning based vehicle re-identification models. For example, the deep joint discriminative learning (DJDL) \cite{djdl} method trains a convolutional neural network to extract discriminative feature representations of vehicle images using identification, verification and triplet loss functions simultaneously. The improved triplet convolutional neural network \cite{imtri} using the classification-oriented loss function and the original triplet loss function is proposed for learning deep feature representations of vehicle images.
Most of the above-mentioned deep learning features \cite{alexnet, vgg, googlenet, djdl, imtri} are holistical, which are learned by a deep learning network ended up with several fully connection layers. Although these deep learning methods make great progresses of vehicle re-identification, they do not design a specific solution for processing viewpoint variations, which is a crucial challenge for vehicle re-identification.

For better handling viewpoint variations, the adversarial bi-directional long short-term memory (LSTM) network (ABLN) is proposed in \cite{abln}. ABLN uses LSTM to model transformations across continuous view variations of a vehicle and adopts the adversarial architecture to enhance training. Thus, a global vehicle representation containing all views' information can be inferred from only one visible view, and then used for learning to measure the distance between two vehicles with arbitrary views. Similar to ABLN, the spatially concatenated convolutional network (SCCN) and CNN-LSTM bi-directional loop (CLBL) are further proposed in \cite{dhmi} to address the challenging of viewpoint variations. However, all of ABLN \cite{abln}, SCCN \cite{dhmi} and CNN-LSTM \cite{dhmi} require a vehicle dataset, where each vehicle has images in densely sampled camera viewpoints. However, this is hard to acquire in practical video surveillance systems. Therefore, there is still ample room for vehicle re-identification by thoroughly considering viewpoint variations.

\subsection{Similarity Metric for Vehicle Re-identification}
 For similarity metrics for vehicle re-identification, FACT \cite{veri2016} applies the Euclidean or Cosine distance between a vehicle pair described with deep learning features to measure the similarity, as done in many face recognition algorithms \cite{deepid,deepid2}. Moreover, NuFACT \cite{veri2017} calculates the similarity of the query and gallery vehicle images measured by the Euclidean distance in the discriminative null space \cite{null}. In addition, DRDL \cite{drdl} proposes a deep relative distance learning method. This method exploits a two-branch convolutional neural network to transform raw vehicle images into an Euclidean space, where the distance can be directly used to measure the similarity of arbitrary two vehicles.

Besides, multi-modal vehicle re-identification methods are also proposed to improve vehicle similarity metrics. For example, the progressive and multi-modal vehicle re-identification (PROVID) \cite{veri2017} is presented to obtain a more accurate vehicle searching. The PROVID method first applies the NuFACT method to make a coarse searching. Then, it makes a fine searching based on a vehicle license plate verification model so that the the re-identification accuracy is improved. In addition, the two-stage framework (i.e., siamese convolutional neural network (Siamese-CNN) and path long short-term memory (Path-LSTM) network) \cite{vst} incorporates complex
 spatial-temporal information for effectively regularizing the vehicle re-identification results. However, it is obvious that those multi-modal vehicle re-identification methods require the extra vehicle information (e.g., license plate, spatial-temporal information) and thus the additional computational load.

Based on the above analysis, vehicle re-identification method could improve the performance by resisting the viewpoint variation. Furthermore, from the viewpoint of practical applications, the vehicle re-identification method could not require extra vehicle information and introduce extra computational load. Motivated by these, the main
objective of our proposed approach lies in: how to
develop a more effective deep learning feature representation method for vehicle re-identification, without using extra vehicle information.
For that, we propose quadruple directional deep learning features (QD-DLF) for vehicle re-identification in this paper.
The most significant contribution of this paper is to fuse quadruple directional deep features learned by using quadruple directional average pooling layers, which constructs viewpoint robust features to significantly boost the vehicle re-identification performance. The details of the proposed QD-DLF will be described in the following Section.

\begin{figure*}[tp]
    \centering
    \includegraphics[width=1.0\linewidth]{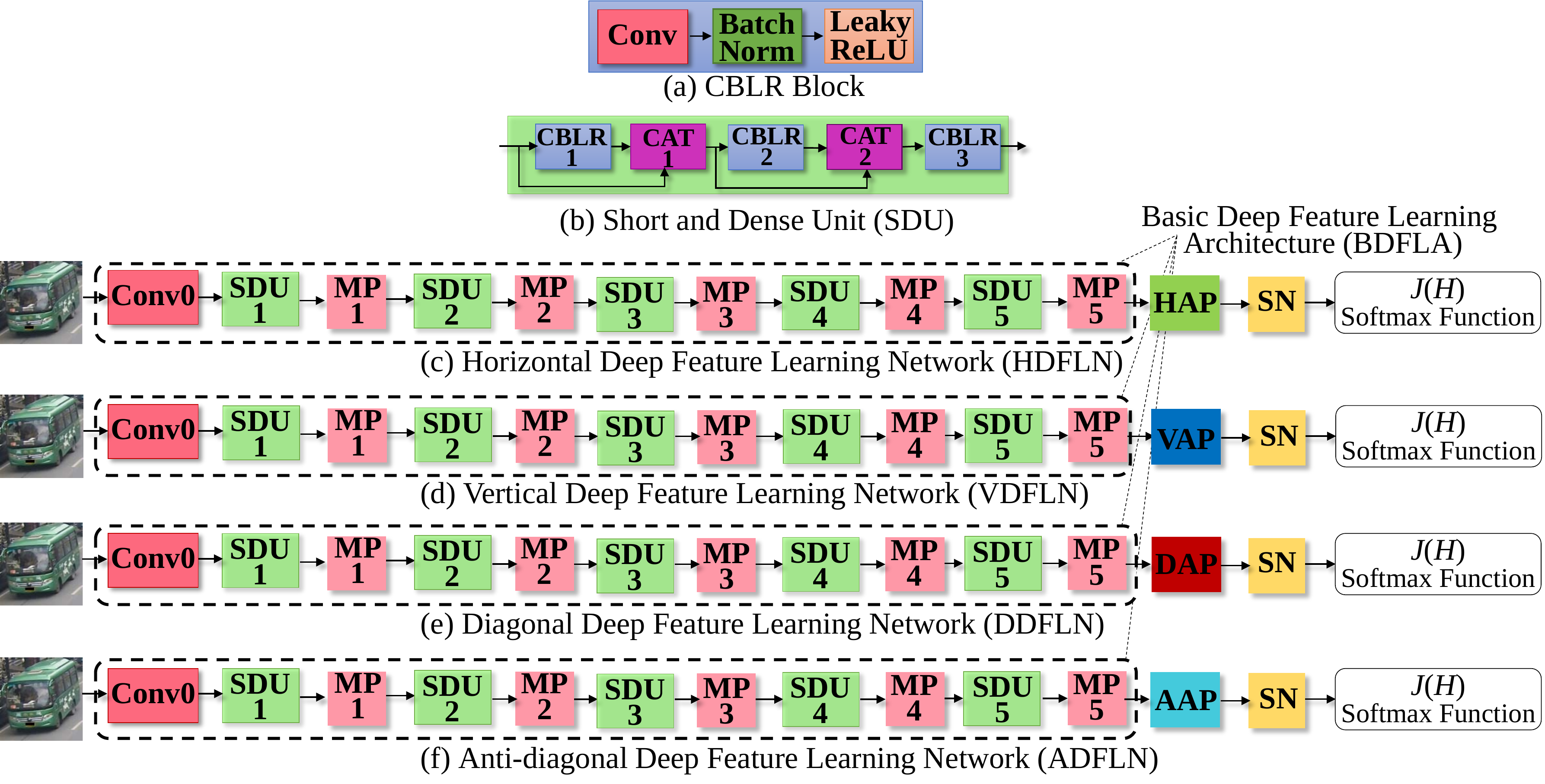}
    \caption{The diagrams of the proposed quadruple directional deep feature learning networks. Here, MP, HAP, VAP, DAP, AAP and SN represents max pooling, horizontal average pooling, vertical average pooling, diagonal average pooling, anti-diagonal average pooling and spatial normalization layers, respectively.}
    \label{fig:framework}
\end{figure*}

\section{Vehicle Re-identification Using Quadruple Directional Deep Learning Features}\label{sec:method}

\subsection{Quadruple Deep Feature Learning Networks}

As shown in Fig. \ref{fig:framework}, the proposed feature learning approach consists of quadruple directional (i.e., horizontal, vertical, diagonal and anti-diagonal) deep feature learning networks (i.e., HDFLN, VDFLN, DDFLN, ADFLN). More specifically, each directional deep feature learning network is composed of a common basic deep feature learning architecture (BDFLA), the corresponding directional average pooling layer and a spatial normalization (SN) layer \cite{matcovnnet}.

\subsubsection{\textbf{Basic Deep Feature Learning Architecture}}
 It can be seen from Fig. \ref{fig:framework} that the basic deep feature learning architecture (BDFLA) is realized by a shortly and densely connected convolutional neural network, which is constructed by a list of short and dense units (SDUs) and max pooling layers. For convenient description, convolutional, batch normalization \cite{bnorm} and Leaky ReLU \cite{leakrelu} layers are sequently packaged to construct a CBLR block, as shown in Fig. \ref{fig:framework} (a). Then, three CBLR blocks are densely connected with two concatenation layers (i.e., CAT1 and CAT2) to build a short and dense unit (SDU), as shown in Fig. \ref{fig:framework} (b).
 Each concatenation layer concatenates the input images according to the channel dimension. Finally, one convolutional layer (i.e., Conv0), a batch normalization, five SDUs (i.e., SDU1-SDU5), and five max pooling layers (i.e., MP1-MP5) are packaged in turn to obtain the basic deep feature learning architecture.

\subsubsection{\textbf{Quadruple Directional Average Pooling Layers}}
To comprehensively describe vehicle images from different directions, quadruple directional (i.e., horizontal, vertical, diagonal and anti-diagonal) average pooling layers are designed.
Assume that the basic feature maps produced by the BDFLA is $X \in \Re{^{d \times d \times c}}$, where $d$ and $c$ represent the height/width, and channel sizes, respectively. The developed quadruple directional average pooling layers can be described as follows.

\begin{figure}[t]
    \centering
    \includegraphics[width=.8\linewidth]{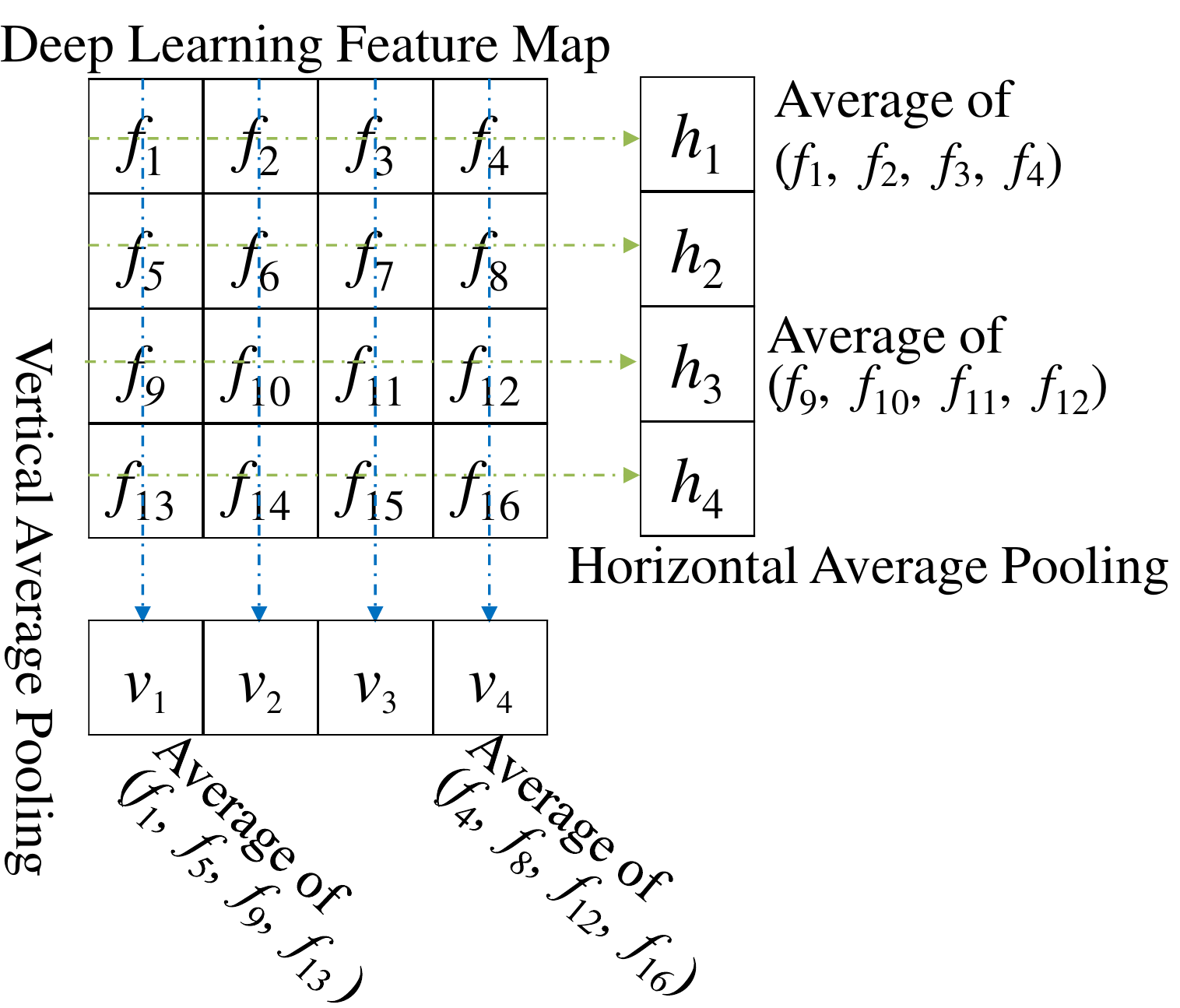}
        \vspace{-.4cm}
    \caption{The schematic diagram of horizontal and vertical average pooling operations.}
    \vspace{-.4cm}
    \label{fig:hvpool}
\end{figure}

\begin{figure}[t]
    \centering
    \includegraphics[width=1\linewidth]{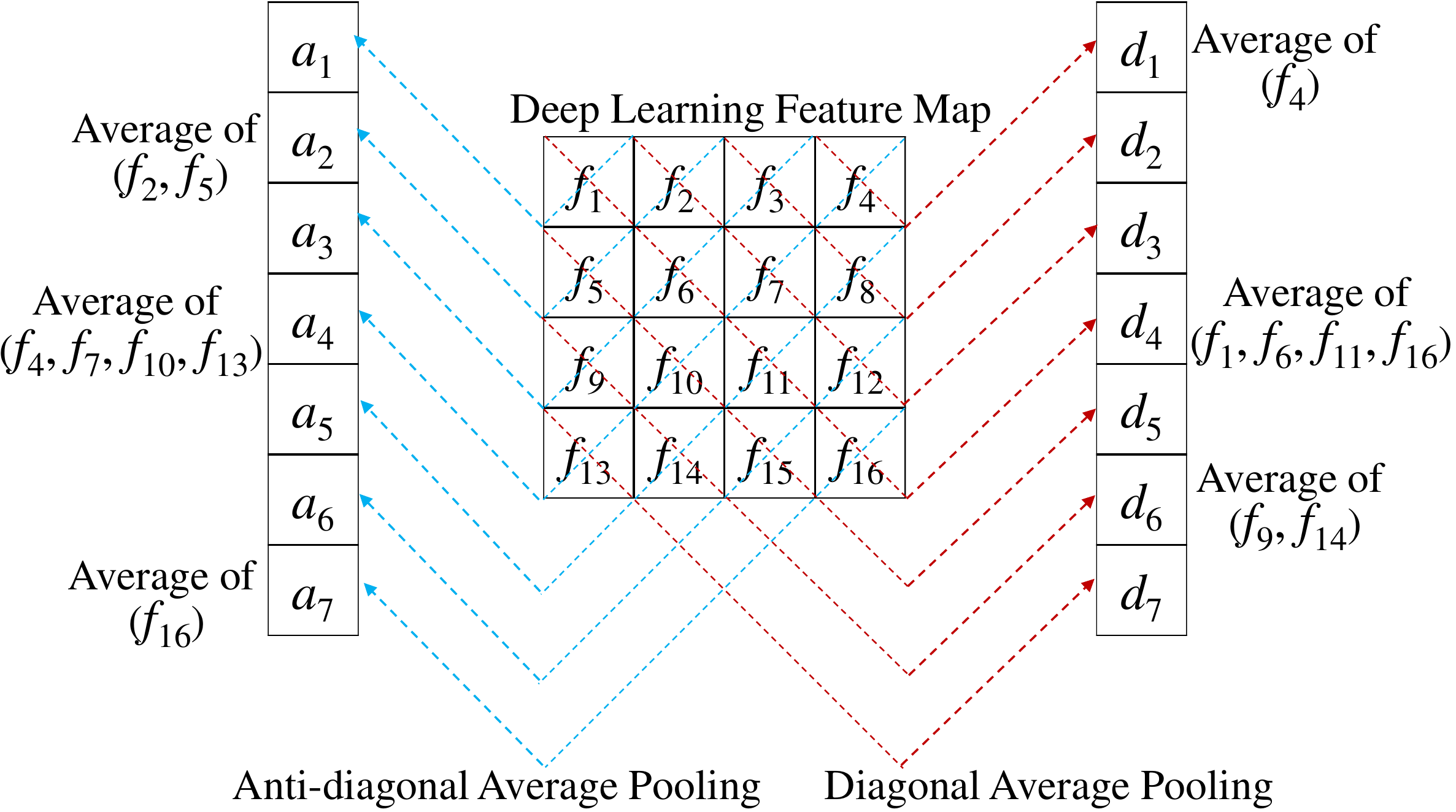}
    \caption{The schematic diagram of diagonal and anti-diagonal average pooling operations.}
    \vspace{-.4cm}
    \label{fig:dapool}
\end{figure}

\begin{enumerate}
   \item [i)]{\textbf{Horizontal Average Pooling (HAP) Layer}: The HAP layer averages each row of $X$ into a single point to obtain the horizontal average pooling feature map $P\in \Re{^{d \times 1 \times c}}$, as shown in Fig. \ref{fig:hvpool}. For example, $h_1$ is equal to the average of $f_1$, $f_2$, $f_3$ and $f_4$, that is, $h_1=\frac{1}{4}(f_1+f_2+f_3+f_4)$}.
   \item [ii)]{\textbf{Vertical Average Pooling (VAP) Layer}:
   The VAP layer averages each column of $X \in \Re{^{d \times d \times c}}$ into a single point to obtain the vertical average pooling feature $Q \in \Re{^{1 \times d \times c}}$, as shown in Fig. \ref{fig:hvpool}. Note that $Q$ is transposed into $ Q_{t} \in \Re{^{d \times 1 \times c}} $ in a practical testing process to make the dimension of $Q_t$ be compatible with that of $P$ (i.e., the output of the HAP layer). For instance, $v_4$ is equal to the average of $f_4$, $f_8$, $f_{12}$ and $f_{16}$, that is, $v_1=\frac{1}{4}(f_4+f_8+f_{12}+f_{16})$}, as show in Fig. \ref{fig:hvpool}.

   \item [iii)]{\textbf{Diagonal Average Pooling (DAP) Layer}:
   The DAP layer averages multiple elements of the feature map $X \in \Re{^{d \times d \times c}}$ according to the diagonal direction, as shown in Fig. \ref{fig:dapool}. For example, $d_6$ is equal to the average of $f_9$ and $f_{14}$, that is, $d_6=\frac{1}{2}(f_9+f_{14})$.}
   \item [iv)]{\textbf{Anti-diagonal Average Pooling (AAP) Layer}:
   The AAP layer averages multiple elements of the feature map $X \in \Re{^{d \times d \times c}}$ according to the anti-diagonal direction, as shown in Fig. \ref{fig:dapool}. For instance, $a_4$ is equal to the average of $f_4$, $f_{7}$, $f_{10}$ and $f_{13}$, that is, $a_4=\frac{1}{4}(f_4+f_{7}+f_{10}+f_{13})$.}
 \end{enumerate}

  Since all the above-mentioned HAP, VAP, DAP and AAP layers use a average pooling operation, the forward and backward propagations are briefly introduced as follows. Assume that the input feature map of an average pooling layer is $X=[X_{1}, X_{2}, ..., X_{d}]  \in \Re ^{d \times d}$, the average pooling window size is $d \times 1$, and the output feature map is $Y \in \Re ^{1\times d}=[y_{1}, y_{2}, ..., y_{d}]$. Then, the forward propagation of this average pooling layer is calculated as follows:
\begin{equation}\label{equ:avg}
 y_{i} =\frac{1}{d}\sum\limits_{j = 1}^d {{X_{ij}}},
\end{equation}
 where $y_{i}$ is the $i$-th element of $Y$; $X_{ij}$ is $j$-th element of $i$-th column vector of $X$. According to the chain rule, the backward propagation of this average pooling layer can be calculated as follows:
\begin{equation}\label{equ:elementavg}
\begin{split}
\frac{{\partial J}}{{\partial {X_{i1}}}} = \frac{{\partial J}}{{\partial {y_i}}}\frac{{\partial {y_i}}}{{\partial {X_{i1}}}} = \frac{1}{d}\frac{{\partial J}}{{\partial {y_i}}},\\
\frac{{\partial J}}{{\partial {X_{i2}}}} = \frac{{\partial J}}{{\partial {y_i}}}\frac{{\partial {y_i}}}{{\partial {X_{i2}}}} = \frac{1}{d}\frac{{\partial J}}{{\partial {y_i}}},\\
\rm{\rm{~~~~~~~~~~~~~~~~...~~~~~~~~~~~~~~~~}}\\
\frac{{\partial J}}{{\partial {X_{id}}}} = \frac{{\partial J}}{{\partial {y_i}}}\frac{{\partial {y_i}}}{{\partial {X_{id}}}} = \frac{1}{d}\frac{{\partial J}}{{\partial {y_i}}},
\end{split}
\end{equation}
and the matrix form can be formulated as follows:
\begin{equation}\label{equ:avg_b}
\frac{{\partial J}}{{\partial {X_i}}} = \frac{1}{d}{\left[\frac{{\partial J}}{{\partial {y_i}}},\frac{{\partial J}}{{\partial {y_i}}},{\rm{ }}...,{\rm{ }}\frac{{\partial J}}{{\partial {y_i}}}\right]^{\rm{T}}},
\end{equation}
 where ${X_i} = {\left[ {{X_{i1}},{X_{i2}},...,{X_{id}}} \right]^{\rm{T}}}$ is $i$-th column vector of $X$; $J$ is the objective function (i.e., Eq. (\ref{equ:J})) of the overall learning framework and will be discussed in the following subsection.

\subsubsection{\textbf{Spatial Normalization Layer}}

As shown in Fig. \ref{fig:framework}, a spatial normalization (SN) layer \cite{matcovnnet} is exploited to follow each directional average pooling layer. It is to make each dimension of the directional average pooling feature maps unified distributing in [0, 1), which is beneficial to prevent a specific dimension whose value is too predominate.
Assume that the input of a SN layer is $P \in \Re{^{d \times c}}$. Then, the corresponding output $Z$ of the SN layer can be calculated as follows:

\begin{equation}\label{equ:sn}
 Z^{k}_{j} = \frac{P^{k}_{j}}{\sqrt{1+\sum\nolimits_{l \in {N_j}} {(P^{k}_{l})^2} }},
\end{equation}
where $Z^{k}_{j}$ is $j$-th element of the $k$-th feature map of $Z$, $P^{k}_{j}$ represents the $j$-th element of the $k$-th feature map of $P$, and $N_j$ is the neighborhood size.

The backward propagation of the SN layer can be formulated as follows:
\begin{equation}\label{equ:sn_b}
\footnotesize{
\frac{{\partial J}}{{\partial {P^{k}_{j}}}} = \frac{{\frac{{\partial J}}{{\partial {Z^{k}_{j}}}} - Z^{k}_{j}\sum\nolimits_{l \in {N_j}} {\frac{{\partial J}}{{\partial {Z^{k}_{l}}}}{Z^{k}_{l}}} }}{{\sqrt {1 + \sum\nolimits_{l \in {N_j}} {  (P^{k}_{l})^2  } } }}
 }.
\end{equation}

Based on the above-mentioned basic deep feature learning network, quadruple directional average pooling layers, and SN layer, quadruple deep feature learning networks (QDFLNs) are constructed, consisting of horizontal deep feature learning network (HDFLN), vertical deep feature learning network (VDFLN), diagonal deep feature learning network (DDFLN), and anti-diagonal deep feature learning network (ADFLN), as shown in Fig. \ref{fig:framework}.

 The recommended parameter configuration of the proposed QD-DLF method is listed in Table \ref{tab:parameter}. The channel numbers of Conv0, SDU1, SDU2, SDU3, SDU4 and SDU5 are 64, 64, 128, 192, 256 and 320, respectively. The scope of Leaky ReLU layer of SDU5 is 0, and that of the others is 0.15. The sub-window for Conv0 and SDU represents a filter size. Specifically, for pooling layers (i.e. MP1-MP5), it means a pooling window size, while for the spatial normalization (i.e. SN) layers, it denotes a normalization window size. Conv0 and five SDUs (i.e., SDU1-SDU5) apply $3\times3$ sized filters. Five max pooling layers use $3\times3$ sized pooling windows. Among the quadruple directional average pooling layers (i.e., HAP, VAP, DAP and AAP), the pooling window sizes for HAP and VAP layers are $1\times4$ and $4\times 1$, respectively, while for DAP and AAP the maximum number of pooled elements is 4, as show in Fig. \ref{fig:dapool}. The normalization neighborhood size (i.e., $N_j$ in Eq. (\ref{equ:sn})) in all spatial normalization layers is $4$. Moreover, only those strides working on five MP layers are set as 2 pixels, and those on the others are set as 1 pixel.

\begin{table}[t]
\centering
\renewcommand{\arraystretch}{1.5}
\caption{The parameter configuration of the proposed QDFLNs.}\label{tab:parameter}
\setlength{\tabcolsep}{1.2pt}
\begin{tabular}{c|c|c|c|c|c}
  \hlinewd{1.5pt}
  Name        & Channels
              & \begin{tabular}{c}
                Scope of \\
                Leaky ReLU  \\
                \end{tabular}
              & \begin{tabular}{c}
              Sub-window\\
                  ($h\times w$) \\
                \end{tabular}
              & Stride & Output Size
              \\
\hlinewd{0.8pt}
    Conv0           & 64  & 0.15  &$3\times3$  &1   &$128\times128\times64$     \\
\hline
    SDU1           & 64  & 0.15  &$3\times3$   &1   &$128\times128\times64$    \\
\hline
    MP1            & 64  & -     &$3\times3$   &2   &$64\times64\times64$    \\
\hline
    SDU2           & 128 & 0.15  &$3\times3$   &1   &$64\times64\times128$    \\
\hline
    MP2            & 128 & -     &$3\times3$   &2   &$32\times32\times128$    \\
\hline
    SDU3           & 192 & 0.15  &$3\times3$   &1   &$32\times32\times192$    \\
\hline
    MP3            & 192 & -     &$3\times3$   &2   &$16\times16\times192$    \\
\hline
    SDU4           & 256 & 0.15  &$3\times3$   &1   &$16\times16\times256$    \\
\hline
    MP4            & 256 & -     &$3\times3$   &2   &$8\times8\times256$    \\
\hline
    SDU5           & 320 & 0     &$3\times3$   &1   &$8\times8\times320$    \\
\hline
    MP5            & 320 & -     &$3\times3$   &2   &$4\times4\times320$    \\
\hline
    HAP            & 320 & -     &$1\times4$   &1   &$4\times1\times320$     \\
\hline
    VAP            & 320 & -     &$4\times1$   &1   &$1\times4\times320$     \\
\hline
    DAP            & 320 & -     &$4$   &1   &$7\times1\times320$     \\
\hline
    AAP            & 320 & -     &$4$   &1   &$7\times1\times320$     \\
\hlinewd{1.5pt}
\end{tabular}
\vspace{-.3cm}
\end{table}

\subsection{Objective Function}

Similar to \cite{deepid, hogcnn}, the softmax function is utilized to build the objective function of the proposed method, as follows:
\begin{equation}\label{equ:J}
\small
J{(W)} =  \frac{1}{K}[{\sum\limits_{k = 1}^K {\sum\limits_{c = 1}^C {{\ell}({y^{(k)}}= c) \log \frac{{{e^{{W_c}^T{X^{(k)}}}}}}{{\sum\nolimits_{p = 1}^C {{e^{{W_p}^T{X^{(k)}}}}} }}} } }]+\frac{1}{2} \alpha \left\| W \right\|_2^2,
\end{equation}
 where $W=[W_1, W_2, ..., W_C] \in {\Re ^{d\times C}}$ is the projection matrix used to predicate a vehicle's class label, $X^{(k)}$ is the deep learning feature of $k$-th training sample, $y^{(k)} \in \{1,2,3,...,C\}$ is the corresponding class label, $\alpha$ is a constant used to control the contribution of the $L_2$ regularization item, $K$ and $C$ represent the numbers of the training samples and classes, respectively, and ${\ell}(\cdot)$ is an indicator function.

\section{Experiment and Analysis}\label{sec:exp}
 To validate the superiority of the proposed quadruple directional deep learning feature (QD-DLF) approach, the performance comparison with multiple state-of-the-art methods is conducted on two challenging databases, namely, VeRi \cite{veri2017} and VehicleID \cite{drdl}. In our experiments, the Euclidean distance is employed to measure the similarity of a vehicle pair described with quadruple deep learning features.
 And two commonly used criteria in the re-identification field, i.e., {cumulative match curve} (CMC) \cite{viper,hybrid} and {mean average precision} (MAP) \cite{market,embed}, are used to evaluate the performance. The CMC shows the identification accuracy rates that a query identity appears in different sized candidate lists. The MAP is used to evaluate the overall performance. For each query, the area under the precision-recall curve is calculated, which is known as average precision (AP). Then, the mean value of APs of all queries is calculated as MAP, which considers both precision and recall of a re-identification method, and thus provides a more comprehensive performance evaluation.

\begin{figure*}[tp]
{
\subfigure[]{
\begin{minipage}[t]{0.33\linewidth}
\includegraphics[width=1.0\linewidth]{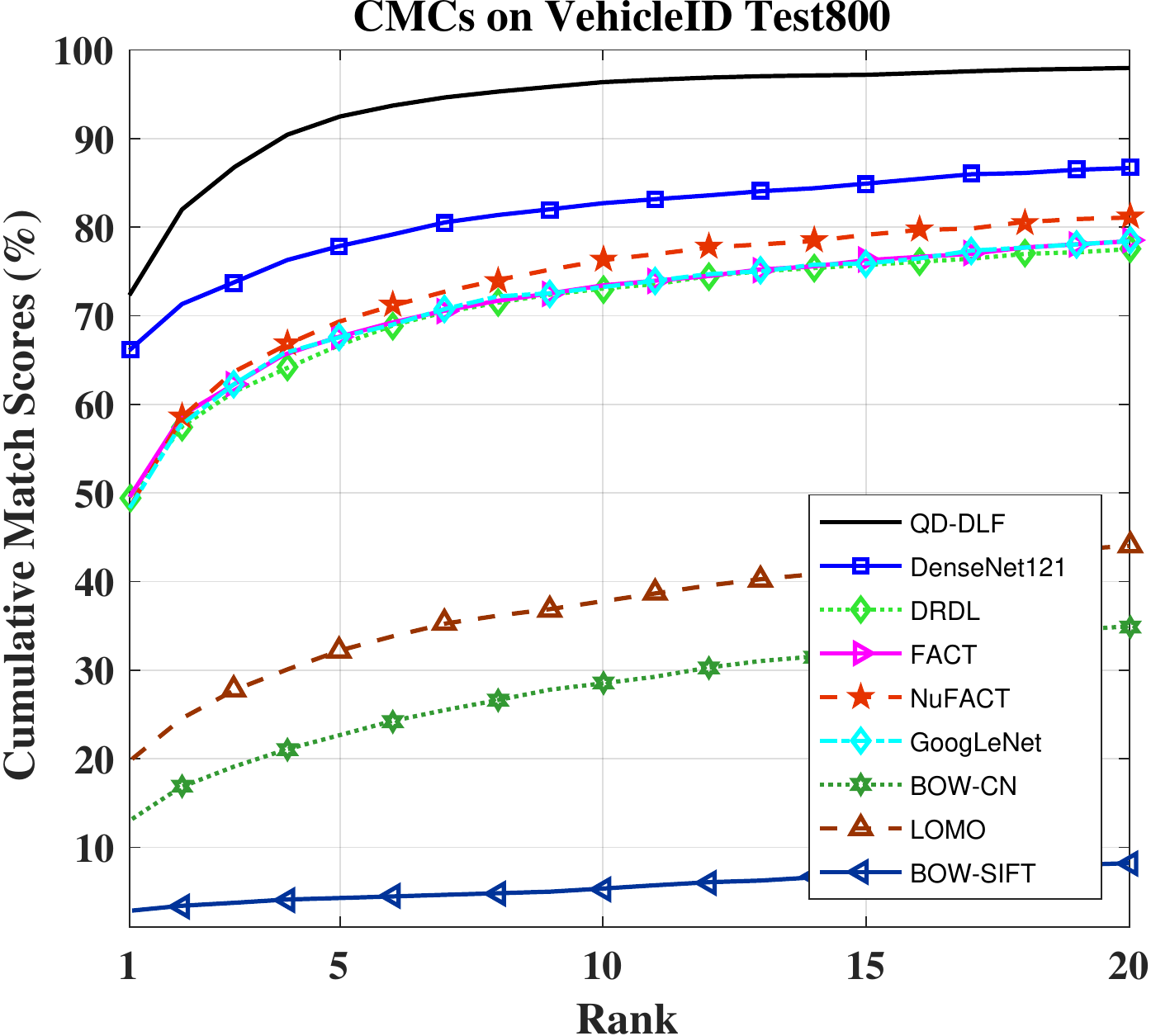}
\label{fig:vehicleid_result:a}
\vspace{-1cm}
\end{minipage}
}
\subfigure[]{
\begin{minipage}[t]{0.33\linewidth}
\includegraphics[width=1.0\linewidth]{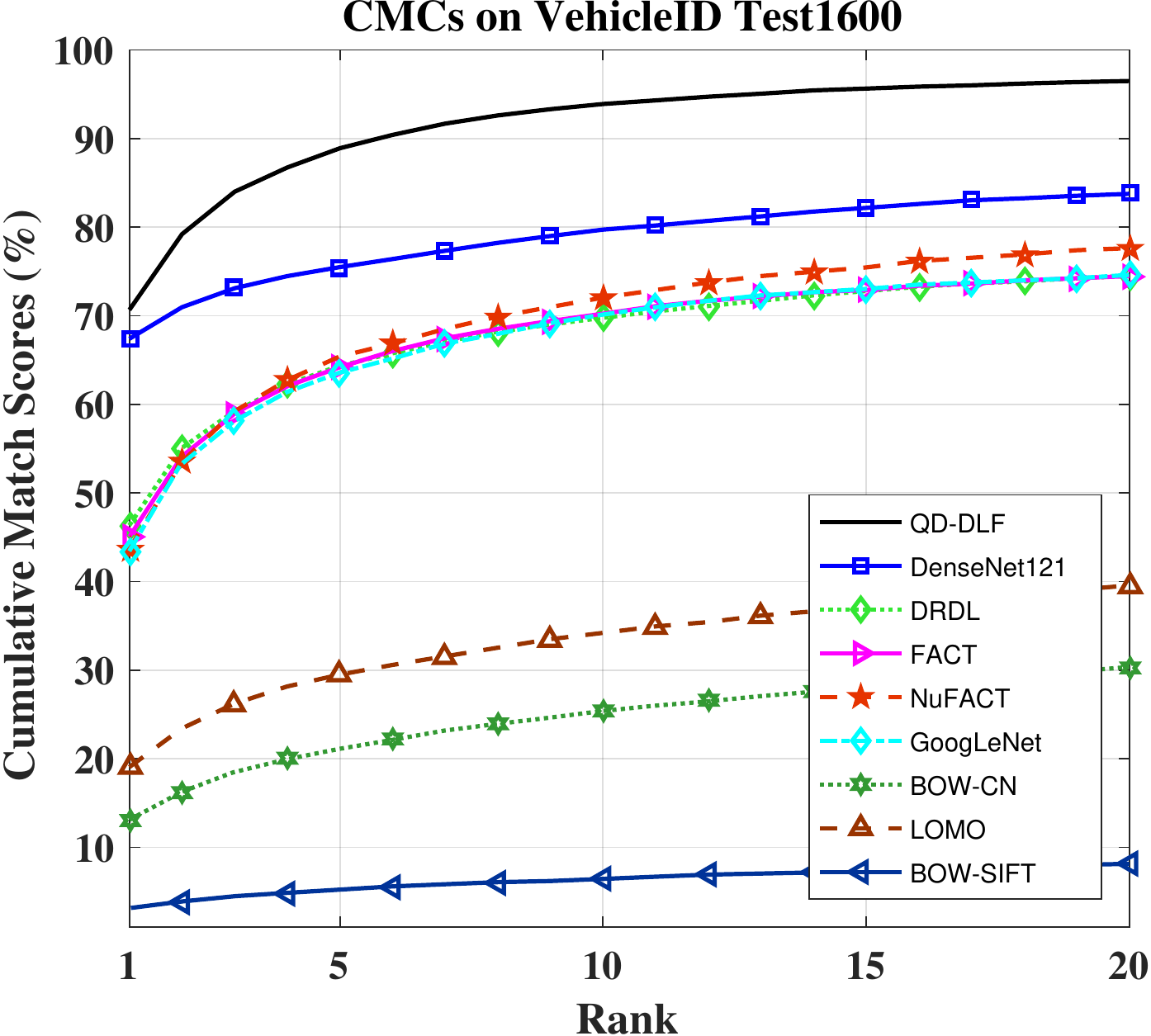}
\label{fig:vehicleid_result:b}
\vspace{-1cm}
\end{minipage}
}
\subfigure[]{
\begin{minipage}[t]{0.33\linewidth}
\includegraphics[width=1.0\linewidth]{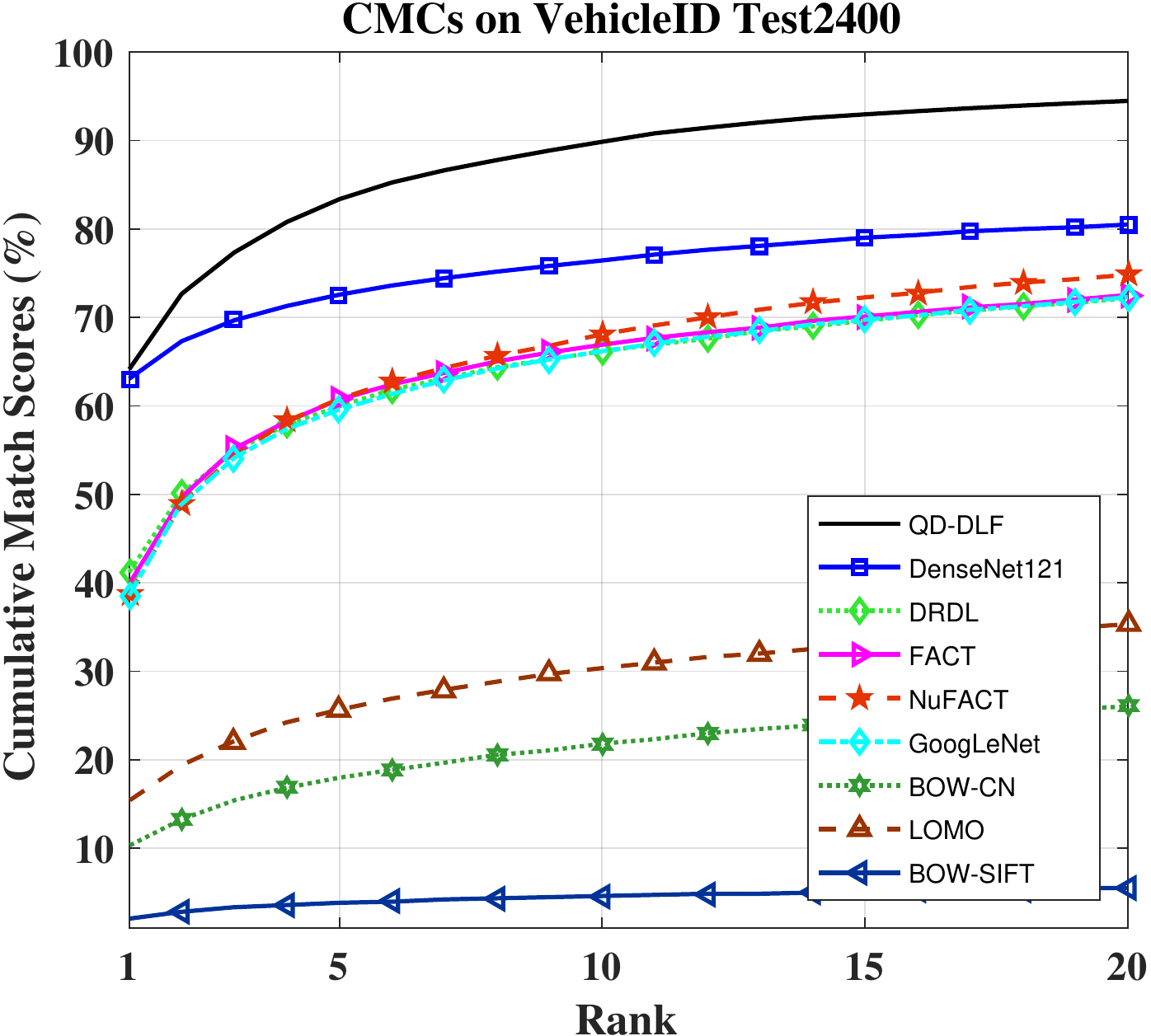}
\label{fig:vehicleid_result:c}
\vspace{-1cm}
\end{minipage}
}
}
\vspace{-.5cm}
\caption{The CMC curve comparisons of the proposed QD-DLF method and multiple state-of-the-art methods on (a) Test800, (b) Test1600, and (c) Test2400 of VehicleID, respectively.} \label{fig:vehicleid_result-overall}
\end{figure*}

\subsection{Training Configuration}
In our experiments, the software tools are Matconvnet \cite{matcovnnet}, CUDA 8.0, CUDNN V5.1, MATLAB 2014 and Visual Studio 2012. The hardware device is workstation configured with a Intel Xeon E3-1505 M v5 CPU @2.80 GHz, a NVIDIA Titan X GPU and 128 GB DDR3 Memory. Moreover, the training settings similar to \cite{hybrid, mytri} are adopted and summarized as follows. All images in these two databases are scaled to $128\times128$ pixels, and each image is further augmented by the horizontal mirror and randomly rotating operations. The randomly rotating operation is applied to randomly rotate an image in ranges $[-3^{\circ}, 0^{\circ}]$ and $[0^{\circ}, 3^{\circ}]$. The weights in each layer are initialized based on a normal distribution $N(0, 0.01)$, and the biases are initialized to 0. The $L_2$ regularization weights $\alpha$ in Eq. (\ref{equ:J}) is set as 0.005 on the VeRi \cite{veri2017} database, while for the larger VehicleID \cite{drdl} database, it is set as 0.001. The size of mini-batch is 128 including 64 positive and 64 negative image pairs, and both positive and negative pairs are randomly selected from the whole database. The momentums are set to 0.9. The learning rates start with 0.01 and are gradually decreased along the training progress. That is, if the objective function is convergent at a stage, the learning rates are reduced to 1/10 of the current values, and the minimum learning rates are 0.001.

\subsection{Databases}
 \textbf{\emph{VeRi}}~\cite{veri2017} is captured by 20 cameras in unconstrained traffic scenarios and each vehicle is captured by 2-18 cameras under different viewpoints, illuminations, occlusions and resolutions. The VeRi dataset is divided into a training subset containing 37,781 images of 576 subjects and a testing subset with 13,257 images of 200 subjects. For the evaluation, one image of each vehicle captured from each camera is applied as a query, then a query set containing 1,678 images of 200 subjects and a gallery including 11,579 image of 200 subjects are finally obtained. Furthermore, only the cross-camera vehicle re-identification is evaluated, which means that if a probe image and a gallery image are captured under the same camera viewpoint, the corresponding matching result will be excluded in the final performance evaluation.

 \textbf{\emph{VehicleID}}~\cite{drdl} is captured in daytime by multiple real-world surveillance cameras distributed in a small city of China. There are 221,763 images of 26,267 subjects in the entire database. Each vehicle is captured from either a front viewpoint or a back viewpoint. The training subset consists of 110,178 images of 13,134 subjects. In addition, VehicleID provides three testing subsets, Test800, Test1600 and Test2400, for evaluating the performance in different data scales. Specifically, Test800 includes 800 gallery images and 6,532 probe images of 800 subjects. Test1600 contains 1600 gallery images and 11,395 probe images of 1,600 subjects. Test2400 consists 2400 gallery images and 17,638 probe images of 2,400 subjects.

\begin{table}[tp]
\centering
\renewcommand{\arraystretch}{1.5}
\caption{The performance (\%) comparison of the proposed QD-DLF and multiple state-of-the-art methods on VeRi.}\label{tab:veri}
\setlength{\tabcolsep}{0.8pt}
\begin{tabular}{c|c|c|c}
\hlinewd{1.5pt}
Methods                             &MAP     &Rank=1       &Rank=5\\
\hlinewd{0.8pt}
Proposed QD-DLF                              &\textbf{61.83}& \textbf{88.50}& 94.46\\ 
\hline
Siamese-CNN+Path-LSTM \cite{vst}              &58.27 &83.49 &90.04\\
\hline
PROVID \cite{veri2017}              &{53.42}   &81.56        &\textbf{95.11}\\
\hline
NuFACT + Plate-SNN \cite{veri2017}  &50.87   &81.11        &92.79\\
\hline
NuFACT + Plate-REC \cite{veri2017}  &48.55   &76.88        &91.42\\
\hline
NuFACT \cite{veri2017}              &48.47   &76.76        &91.42\\
\hline
DenseNet121 \cite{dense}            &45.06   &80.27        &91.12 \\
\hline
SCCN-Ft+CLBL-8-Ft \cite{dhmi}       &25.12   &60.83        &78.55\\
\hline
ABLN-Ft-16 \cite{abln}              &24.92   &60.49        &77.33\\
\hline
FACT \cite{veri2016}                &18.75   &52.21        &72.88\\
\hline
GoogLeNet \cite{largecar}           &17.89   &52.32        &72.17\\
\hline
VGG-CNN-M-1024 \cite{drdl}          &12.76   &44.10        &62.63\\
\hline
BOW-CN \cite{market}               &12.20   &33.91        &53.69\\
\hline
LOMO \cite{lomo}                    &9.64    &25.33        &46.48\\
\hline
BOW-SFIT \cite{bowsift}             &1.51    &1.91         &4.53\\
\hlinewd{1.5pt}
\end{tabular}
\vspace{-.5cm}
\end{table}

\subsection{Performance Evaluation}
\subsubsection{{Comparison on VeRi}}
  The performance comparison of the proposed QD-DLF and multiple state-of-art methods on the VeRi database is shown in Table \ref{tab:veri}. It can be found that the proposed QD-DLF acquires the highest MAP (i.e., 61.83\%) and rank-1 identification rate (i.e., 88.50\%) among all methods under comparisons. More details are analyzed as follows.

  Firstly, compared with four multi-modal deep learning based vehicle re-identification methods (i.e., NuFACT + Plate-SNN \cite{veri2017}, NuFACT + Plate-REC \cite{veri2017}, PROVID \cite{veri2017} and Siamese-CNN+Path-LSTM \cite{vst}), the proposed QD-DLF method consistently defeats NuFACT + Plate-SNN \cite{veri2017}, NuFACT + Plate-REC \cite{veri2017} and Siamese-CNN+Path-LSTM \cite{vst} by higher MAPs, rank-1 identification rates and rank-5 identification rates. Although the rank-5 identification rate of the proposed QD-DLF method is a bit lower than that of PROVID \cite{veri2017}, the proposed QD-DLF method achieves much higher MAP and rank-1 identification rate, and thus is still superior to PROVID \cite{veri2017}.

  Secondly, compared with those single modal deep learning based vehicle re-identification methods (i.e., NuFACT \cite{veri2017}, DenseNet121 \cite{dense}, SCCN-Ft+CLBL-8-Ft \cite{dhmi}, ABLN-Ft-16 \cite{abln}, FACT \cite{veri2016}, GoogLeNet \cite{largecar} and VGG-CNN-M-1024 \cite{drdl}), the proposed QD-DLF methods shows a larger accuracy improvement. Specifically, the best single modal deep learning based vehicle re-identification method, i.e., NuFACT \cite{veri2017}, only obtains a 48.47\% MAP, a 76.76\% rank-1 identification rate and a 91.42\% rank-5 identification rate, which are much lower than those of the proposed QD-DLF method. Moreover, it can be seen that SCCN-Ft+CLBL-8-Ft \cite{dhmi} and ABLN-Ft-16 \cite{abln} do not obviously show the superiority on the VeRi database, although they specially consider the viewpoint variation. This is because each vehicle in the VeRi database is not densely captured by different camera viewpoints, which does not fully meet the requirement of the training data for CCN-Ft+CLBL-8-Ft \cite{dhmi} and ABLN-Ft-16 \cite{abln}, limiting the performance improvement.

  Thirdly, the proposed QD-DLF also obtains much better performance than those hand-crafted feature representation methods, i.e., BOW-CN \cite{market}, LOMO \cite{lomo} and BOW-SFIT \cite{bowsift}.

\begin{table*}[tp]
\centering
\renewcommand{\arraystretch}{1.5}
\caption{The performance (\%) comparison of the proposed QD-DLF and multiple state-of-the-art methods on VehicleID.}\label{tab:vehicleid_result}
\setlength{\tabcolsep}{0.8pt}
\begin{tabular}{c ||c|c|c|| c|c|c||  c|c|c||   c|c|c}
\hlinewd{1.5pt}
 \multirow{2}{*}{Method} &\multicolumn{3}{c||} {Test800}       & \multicolumn{3}{c||} {Test1600}    & \multicolumn{3}{c||}{Test2400}     & \multicolumn{3}{c} {Average}\\
\cline{2-13}
                            &MAP     & Rank=1      & Rank=5       &MAP     & Rank=1      & Rank=5      &MAP    & Rank=1       & Rank=5      &MAP     & Rank=1      &Rank=5 \\
\hlinewd{0.8pt}
Proposed QD-DLF &\textbf{76.54}& \textbf{72.32}& \textbf{92.48}         &\textbf{74.63}&\textbf{70.66}&\textbf{88.90}        &\textbf{68.41}&\textbf{64.14}&\textbf{83.37}        &\textbf{73.19}   &\textbf{69.04}        &\textbf{88.25}     \\
\hline
 DJDL \cite{djdl}                  &N/A   &72.3   &85.7 &N/A  &70.8 &81.8 &N/A  &68.0 &78.9 &N/A  &70.4 &82.1\\
\hline
 DenseNet121 \cite{dense}   &68.85   &{66.10}        &77.87         &{69.45}   &{67.39}        &75.49        &{65.37}   &{63.07}        &72.57        &{67.89}   &{65.52}        &75.31     \\
\hline
 Improved Triplet CNN \cite{imtri} &N/A   &69.9   &87.3 &N/A  &66.2 &82.3 &N/A  &63.2 &79.4 &N/A  &66.4 &83.0\\
 \hline
 DRDL \cite{drdl}       &N/A     &48.91        &66.71         &N/A     &46.36        & 64.38       &N/A     &40.97        &60.02        &N/A     &45.41        &63.70     \\
\hline
 FACT \cite{veri2016}       &N/A     &49.53        &67.96         &N/A     &44.63        & 64.19       &N/A     &39.91        &60.49        &N/A     &44.69        &64.21     \\
\hline
 NuFACT \cite{veri2017}     &N/A     &48.90        &69.51         &N/A     &43.64        & 65.34       &N/A     &38.63        &60.72        &N/A     &43.72        &65.19     \\
\hline
 GoogLeNet \cite{largecar} &N/A     &47.90        &67.43         &N/A     &43.45        & 63.53       &N/A     &38.24        &59.51        &N/A     &43.20        &60.04     \\
\hline
 LOMO \cite{lomo}           &N/A     &19.74        &32.14         &N/A     &18.95        &29.46        &N/A     &15.26        &25.63        &N/A     &17.98        &3.76      \\
\hline
 BOW-CN \cite{market}       &N/A     &13.14        &22.69         &N/A     &12.94        & 21.09       &N/A     &10.20        &17.89        &N/A     &12.09        &20.56     \\
\hline
 BOW-SIFT \cite{bowsift}    &N/A     &2.81         &4.23          &N/A     &3.11         &5.22         &N/A     &2.11         &3.76         &N/A     &2.68         &3.76      \\
\hlinewd{1.5pt}
\end{tabular}
\end{table*}

\subsubsection{{Comparison on VehicleID}}
The performance comparisons of the proposed QD-DLF and multiple state-of-art methods on VehicleID database are shown in Fig. \ref{fig:vehicleid_result-overall} and Table \ref{tab:vehicleid_result}. Firstly, it can be observed that deep learning based methods (i.e., DJDL \cite{djdl}, DenseNet121 \cite{dense}, Improved Triplet CNN \cite{imtri}, DRDL \cite{drdl}, FACT \cite{veri2016}, NuFACT \cite{veri2017} and GoogLeNet \cite{largecar}) obviously defeat traditional methods (i.e., LOMO \cite{lomo}, BOW-CN \cite{market} and BOW-SIFT \cite{bowsift}) on this large scale database. Secondly, the proposed QD-DLF method outperforms all deep learning based methods under comparison, including DJDL \cite{djdl}, DenseNet121 \cite{dense}, Improved Triplet CNN \cite{imtri}, DRDL \cite{drdl}, FACT \cite{veri2016}, NuFACT \cite{veri2017} and GoogLeNet \cite{largecar}, on the Test800, Test1600 and Test2400 subsets of the VehicleID database.

\begin{figure}[tp]
{
\subfigure[]{
\begin{minipage}[t]{0.9\linewidth}
\includegraphics[width=0.9\linewidth]{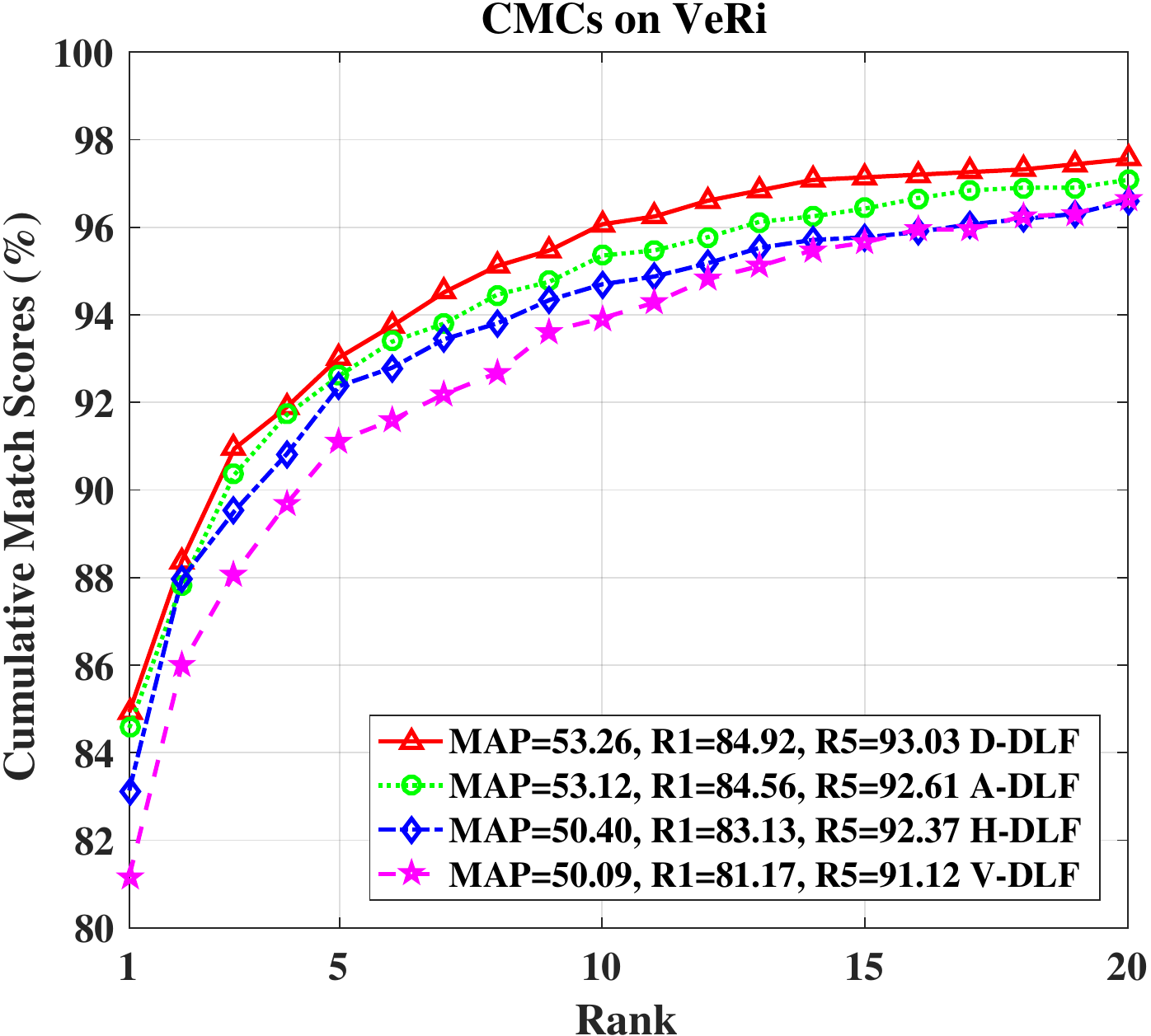}
\label{fig:vehicleid_result:b}
\vspace{-1cm}
\end{minipage}
}
\subfigure[]{
\begin{minipage}[t]{0.9\linewidth}
\includegraphics[width=0.9\linewidth]{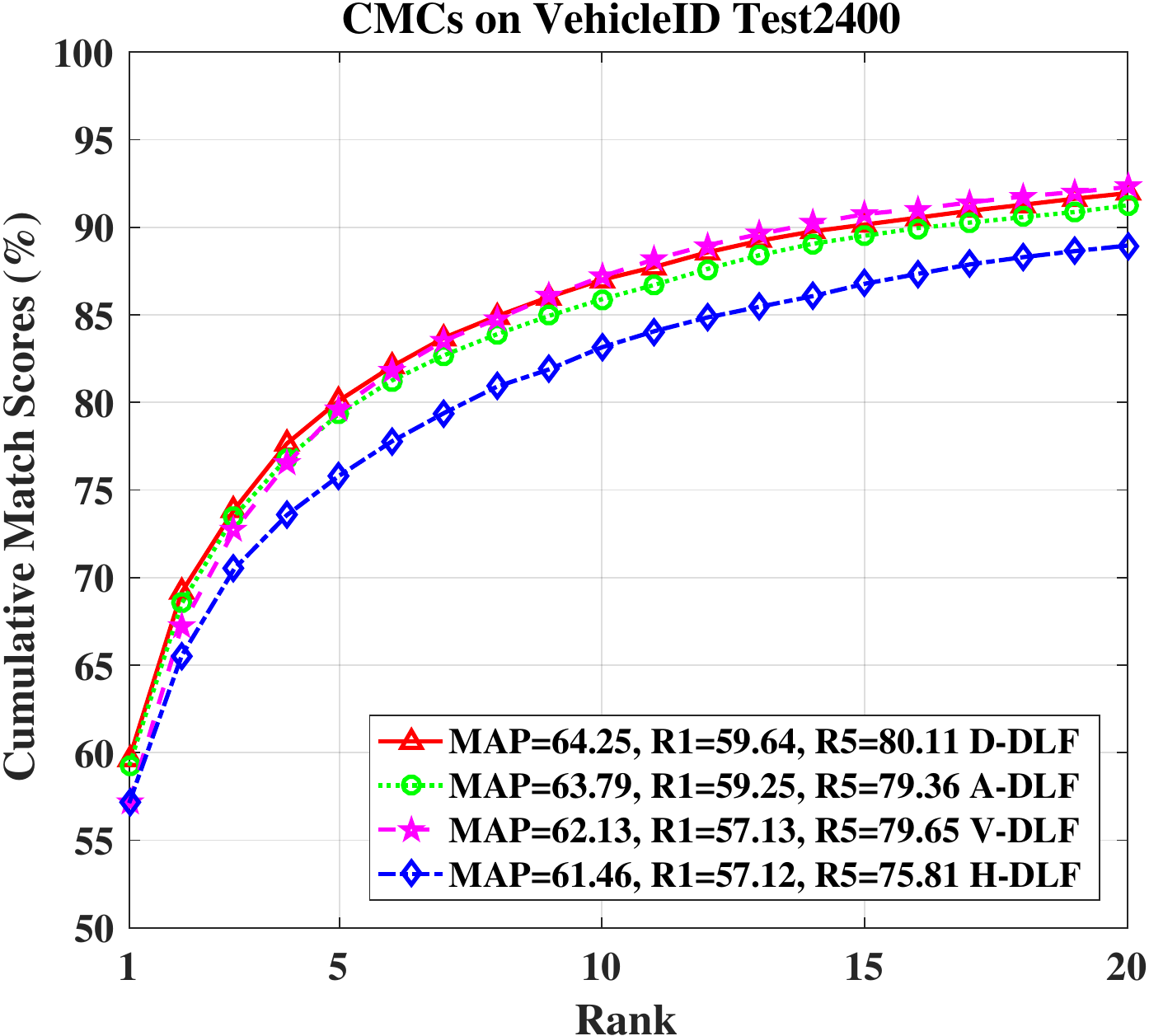}
\label{fig:vehicleid_result:c}
\vspace{-1cm}
\end{minipage}
}
\caption{The performance (\%) comparison of diagonal deep learning feature (D-DLF), anti-diagonal deep learning feature (A-DLF), horizonal deep learning feature (H-DLF) and veridical deep learning feature (V-DLF) on (a) VeRi, (b) Test2400 of VehicleID, respectively.} \label{fig:best_direction}
}
\end{figure}

\subsubsection{Comparison on Each Directional Deep Learning Feature}
In addition, we further comprehensively analyze the performance resulted from each directional deep learning feature. The corresponding results on VeRi and VehicleID databases are shown in Fig. \ref{fig:best_direction}, and Table \ref{tab:role_fusion}, where horizontal, vertical, diagonal and anti-diagonal deep learning features are denoted as H-DLF, V-DLF, D-DLF and A-DLF, respectively. The analyses are described in detailed as follows.
 \begin{enumerate}
   \item [i)]{\textbf{Which directional deep learning feature is better for vehicle re-identification?}

   From Fig. \ref{fig:best_direction}, it can be found that the D-DLF/A-DLF defeats H-DLF/V-DLF. It means that the diagonal/anti-diagonal directional deep learning feature is more appropriate than the horizontal/vertical directional deep learning feature for vehicle re-identification. This may be attributed to two aspects. Firstly, although vehicle images are captured from different camera viewpoints, most vehicle images still have a prominent symmetry. Secondly, for the vehicle images with symmetry, the average pooling corresponding to the diagonal or anti-diagonal direction is able to involve more diverse local feature regions to construct more effective features so as to obtain a better performance of vehicle re-identification.
   }

   \item [ii)]{\textbf{Is there contribution from the fusion of various directional deep learning features?}

   By starting from the D-DLF as an example, we gradually fuse more directional deep learning features to see their performance. Here, DA-DLF means the fusion of D-DLF and A-DLF, DAH-DLF means the fusion of D-DLF, A-DLF and H-DLF. Note that the proposed QD-DLF means the fusion of all four directional deep learning features. From Table \ref{tab:role_fusion}, it can be observed that QD-DLF fusing quadruple directional deep learning features obtains the best performance, DAH-DLF defeats DA-DLF and D-DLF, and DA-DLF outperforms D-DLF. It can be found that the more directional deep learning features are, the better the performance is. This study shows that the fusion of various directional deep learning features can indeed contribute to the improvement of the performance.
       }
 \end{enumerate}

\begin{table}[tp]
\centering
\renewcommand{\arraystretch}{1.5}
\caption{The performance (\%) comparison of QD-DLF, DAH-DLF, DA-DLF and D-DLF on VeRi and Test2400 of VehicleID.}\label{tab:role_fusion}
\setlength{\tabcolsep}{1.5pt}
\begin{tabular}{c|c|c|c||c|c|c}
\hlinewd{1.5pt}
\multirow{2}{*}{Methods}   & \multicolumn{3}{c||} {VeRi}     &\multicolumn{3}{c}{Test2400 of VehicleID}    \\
\cline{2-7}
                           &MAP     &Rank=1       &Rank=5    &MAP     &Rank=1       &Rank=5\\
\hlinewd{0.8pt}
QD-DLF       &\textbf{61.83}&\textbf{88.50}& \textbf{94.46}  &\textbf{68.41}&\textbf{64.14}& {83.37}\\
\hline
DAH-DLF      &60.31 &88.62  &94.34                           &68.24 &63.88  &\textbf{83.72} \\
\hline
DA-DLF       &58.16 &87.19  &94.46                           &66.90 &62.53 &82.07\\
\hline
D-DLF        &53.26 &84.92  &93.03                           &64.25 &59.64 &80.11\\
\hlinewd{1.5pt}
\end{tabular}
\vspace{-.3cm}
\end{table}

\begin{table}[tp]
\centering
\renewcommand{\arraystretch}{1.4}
\caption{The performance (\%) comparison of QD-DLF, D-DLF, A-DLF, H-DLF, V-DLF, and F-DLFs on VeRi.}\label{tab:veri_analy}
\setlength{\tabcolsep}{0.8pt}
\begin{tabular}{c|c|c|c}
\hlinewd{1.5pt}
Methods                             &MAP     &Rank=1       &Rank=5\\
\hlinewd{0.8pt}
QD-DLF     &\textbf{61.83}& \textbf{88.50}& \textbf{94.46}\\ 
\hline
D-DLF      &53.26 &84.92 &93.03 \\
\hline
A-DLF      &53.12 &84.56 &92.61 \\
\hline
H-DLF      &50.40 &83.13 &92.37 \\
\hline
V-DLF      &50.09 &81.17 &91.12 \\
\hline
F-DLF-256   &40.99  &80.15          &90.64\\
\hline
F-DLF-512   &39.68  &80.21          &89.87\\
\hline
F-DLF-128   &39.39  &79.68          &91.12\\
\hline
F-DLF-1024  &39.10  &79.86          &89.33\\
\hlinewd{1.5pt}
\end{tabular}
\vspace{-.3cm}
\end{table}

\subsubsection{Comparison on Directional and Deep Holistic Learning Features}
In this work, we also compare the performance resulted from the proposed method with directional deep learning feature and a deep holistic feature learning configuration. This deep holistic feature learning configuration is built by keeping the basic deep feature learning architecture (BDLFA) unchanged and using a \textbf{Full} connection layer instead of one directional pooling layer. The corresponding learned feature is denoted as F-DLF. Since the F-DLF is able to produce holistical features with different dimensions and we do not know the suitable feature dimension in advance, the performances resulted from the holistical features with different dimensions produced by the full connection layer are all obtained. The F-DLF-128, F-DLF-256, F-DLF-512 and F-DLF-1024 represents the corresponding learning configuration that produces 128, 256, 512 and 1024 dimensional holistical features, respectively.

From the results shown in Table \ref{tab:veri_analy}, one can find that among F-DLF-128, F-DLF-256, F-DLF-512 and F-DLF-1024, F-DLF-256 obtains the highest MAP, but is inferior to our weakest directional deep learning feature, V-DLF. Specifically, the MAP of V-DLF is 9.10\% higher than that of F-DLF-256 and the rank-1 identification rate of V-DLF is 1.02\% higher than that of F-DLF-256, respectively. Moreover, the proposed QD-DLF is significant better than F-DLF-256 (i.e., 20.84\% higher MAP and 8.35\% higher rank-1 identification rate).
This study indicates that the proposed directional deep learning features are more suitable and effective than deep holistic learning features to represent vehicle images.

\subsubsection{Running Time Analysis}
 In addition to the accuracy, the efficiency is also very important for vehicle re-identification methods. For that, the running time comparison is conducted in terms of the {feature extraction time} (FET) per image as suggested in the similar person re-identification task \cite{hybrid}. The running time comparison of the proposed QD-DLF method and several state-of-the-art vehicle re-identification methods is shown in Table \ref{tab:time}, where all the methods are implemented in the GPU mode.

\begin{table}[tp]
\centering
\renewcommand{\arraystretch}{1.4}
\caption{The running time comparison of the proposed QD-DLF method and multiple state-of-the-art vehicle re-identification methods. FET represents the feature extraction time.}\label{tab:time}
\setlength{\tabcolsep}{0.8pt}
\begin{tabular}{c|c}
\hlinewd{1.5pt}
Methods     & FET (msec/image)  \\
\hlinewd{0.8pt}
\hline
D-DLF       &2.321\\
\hline
A-DLF       &2.304\\
\hline
H-DLF       &2.296\\
\hline
V-DLF       &2.312\\
\hline
QD-DLF      &11.199\\
\hline
DenseNet121 \cite{dense}   &13.647\\
\hline
GoogLeNet \cite{largecar}            &2.345\\
\hline
VGG-CNN-M-1024 \cite{drdl}           &1.872\\
\hline
\hlinewd{1.5pt}
\end{tabular}
\vspace{-.3cm}
\end{table}

Firstly, one can see that the running time of each proposed directional deep learning feature (i.e., D-DLF, A-DLF, H-DLF, V-DLF) is similar to that of each other. And the FETs of each directional deep learning features are a bit slower than that of VGG-CNN-M-1024 \cite{drdl} and are comparable to that of GoogLeNet \cite{largecar}. Moreover, compared to the ultra-deep model DenseNet121 \cite{dense}, the FET of each directional deep learning feature is about 17\% of that of DenseNet121 \cite{dense}, which demonstrates all of D-DLF, A-DLF, H-DLF, V-DLF are much faster than DenseNet121~\cite{dense}.

Secondly, the running time of the proposed QD-DLF is analyzed. Intuitively, the FET of QD-DLF would be 4 times of that of each proposed single directional deep learning feature. However, from Table \ref{tab:time}, one can find that the FET of QD-DLF is about 5 times of that of each proposed directional deep learning feature. This is due to that the efficiency of simultaneously making the communalization between CPU and GPU for quadruple directional deep learning model is lower than that for each proposed directional deep learning model, since quadruple directional deep learning model is naturally larger than each single directional deep learning model. In addition, it should be pointed out that the FET of proposed QD-DLF is 2.448ms faster than that of DenseNet121~\cite{dense}.

\section{Conclusion}\label{sec:con}
 In this paper, quadruple directional deep learning networks are designed for vehicle re-identification. The quadruple directional deep learning networks are with the same basic deep learning architecture but different directional feature pooling layers. The same basic deep learning architecture is a shortly and densely connected convolutional neural network to extract basic feature maps of an input square vehicle image. After that, a horizontal average pooling (HAP) layer, a vertical average pooling (VAP) layer, a diagonal average pooling (DAP) layer and an anti-diagonal average pooling (AAP) layer are repetitively applied in the proposed quadruple directional deep learning network to compress the basic feature maps into horizontal, vertical, diagonal and anti-diagonal directional feature maps. Finally, these obtained directional feature maps are spatially normalized and concatenated together as a quadruple directional deep learning feature for vehicle re-identification. Through quadruple directional deep learning features learned by the proposed quadruple directional deep learning network, the adverse effect of viewpoint variations is effectively resisted and the performance of vehicle re-identification is thus significantly improved. Extensive experiments on both VeRi and VehicleID databases show that the proposed method is obviously superior to multiple state-of-the-art vehicle re-identification methods.







%
\ifCLASSOPTIONcaptionsoff
  \newpage
\fi

\bibliographystyle{IEEEtran}
{
\bibliography{reffull}
}
\vspace{-1cm}
\begin{IEEEbiography}[{\includegraphics[width=1in,height=1.25in,clip,keepaspectratio]{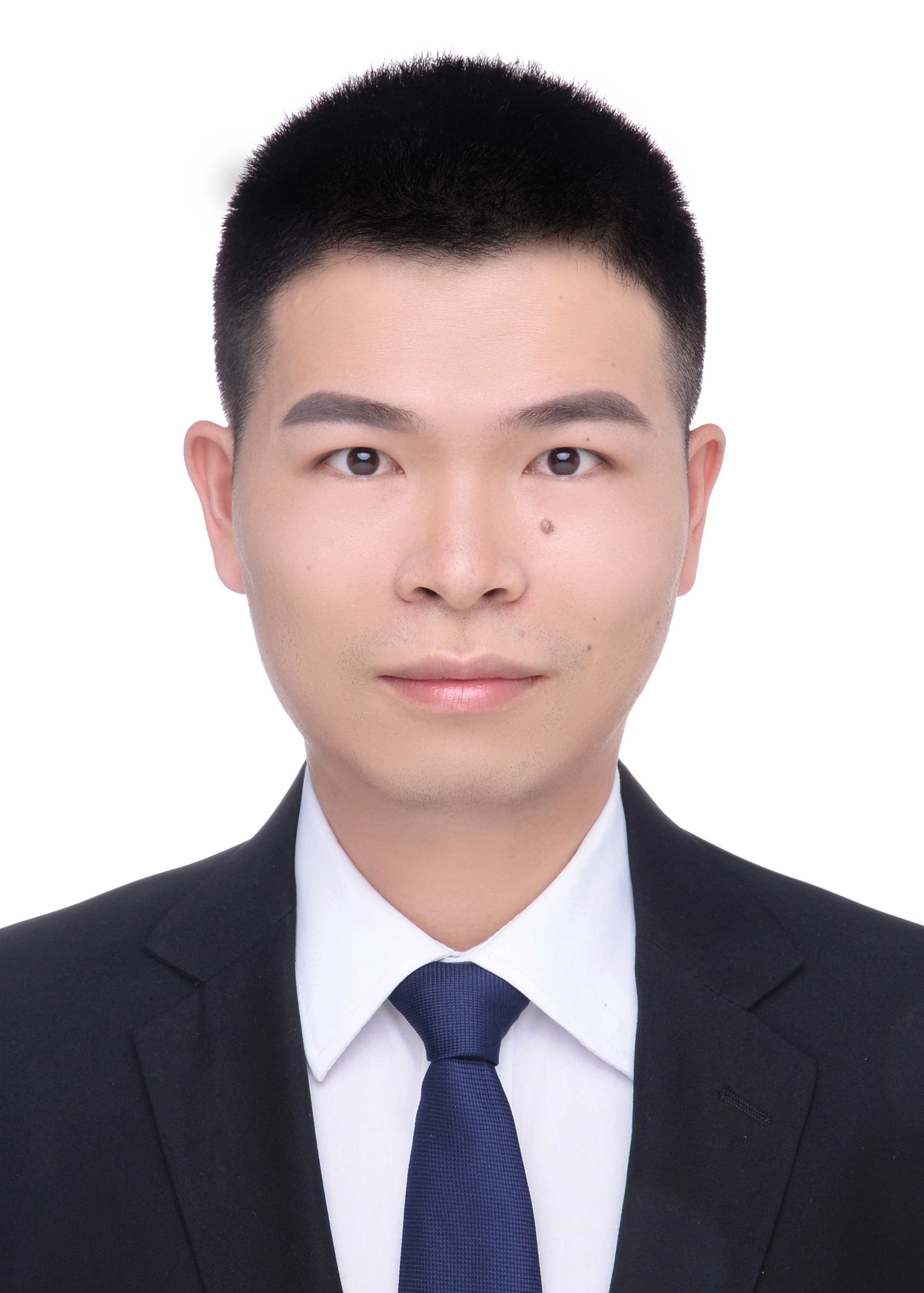}}]{Jianqing Zhu} received the B.S. degree in communication engineering and the M.S. degree in communication and information system from the School of Information Science and Engineering, Huaqiao University, Xiamen, China, in 2009 and 2012, respectively. He received the Ph.D. degree in Institute of Automation, Chinese Academy of Sciences, Beijing, China, in 2015. He is currently an Assistant Professor at the College of Engineering, Huaqiao University, Quanzhou, China. His current research interests include computer vision and pattern recognition, with a focus on image and video analysis, particularly person re-identification, object detection and video surveillance. He was awarded the Best Biometrics Student Paper award at the International Conference on Biometrics in 2015.
\end{IEEEbiography}

\vspace{-1cm}
\begin{IEEEbiography}[{\includegraphics[width=1in,height=1.25in,clip,keepaspectratio]{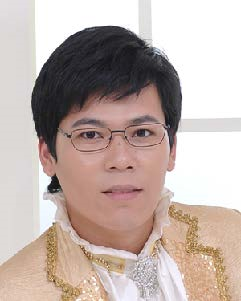}}]{Huanqiang Zeng} (S'10-M'13-SM'18)
received the B.S. and M.S. degrees from Huaqiao University, Xiamen, China and the Ph.D. degree from Nanyang Technological University, Singapore, all in electrical engineering. He was a Postdoctoral Fellow at the Department of Electronic Engineering, The Chinese University of Hong Kong, Hong Kong from 2012 to 2013, and a Research Associate at the Temasek Laboratories, Nanyang Technological University, Singapore in 2008. He is now a Professor at the School of Information Science and Engineering, Huaqiao University, Xiamen, China. His research interests are in the areas of image processing and video coding, machine learning and pattern recognition, and computer vision. He has published more than 80 papers in well-known international journals and conferences. He has been actively serving as the Associate Editor for IET Electronics Letters and International Journal of Image and Graphics, Guest Editor for multiple international journals, including Journal of Visual Communication and Image Representation, Multimedia Tools and Applications, Journal of Ambient Intelligence and Humanized Computing, the General Co-Chair for IEEE International Symposium on Intelligent Signal Processing and Communication Systems 2017 (ISPACS2017), the Technical Program Co-Chair for Asia-Pacific Signal and Information Processing Association Annual Summit and Conference 2017 (APSIPA ASC2017), the Area Chair for IEEE International Conference on Visual Communications and Image Processing (VCIP2015), the Technical Program Committee Member for multiple flagship international conferences. He received the Best Paper Award from Chinese Conference on Signal Processing 2017 (CCSP2017). He is an IEEE Senior member, and a Member of International Steering Committee of International Symposium on Intelligent Signal Processing and Communication Systems.

\end{IEEEbiography}

\begin{IEEEbiography}[{\includegraphics[width=1in,height=1.25in,clip,keepaspectratio]{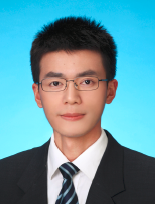}}]{Jingchang Huang} received the Ph.D. degree in Communication and Information Systems from University of Chinese Academic and Sciences, China in 2015. He joined IBM Research - China in 2015 and currently he is a research staff member in Cognitive-IoT team. He mainly focuses on the cognitive data analytics, including but not limited to acoustic, image and time sequences, all of which comes from Internet-of-Thing related applications. His current research interests also include patterns recognition, big data and wireless sensor networks. Dr. Huang has designed and implemented many efficient target detection and classification projects, during which contributes more than 10+ scientific paper in IEEE journals \& Con. and 20+ patent inventions.
\end{IEEEbiography}

\begin{IEEEbiography}[{\includegraphics[width=1in,height=1.25in,clip,keepaspectratio]{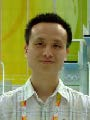}}]{Shengcai Liao} received the B.S. degree in mathematics and applied mathematics from the Sun Yatsen University, Guangzhou, China, in 2005 and the Ph.D. degree from the Institute of Automation, Chinese Academy of Sciences, Beijing, China, in 2010. He was a Post Doctoral Fellow in the Department of Computer Science and Engineering, Michigan State University during 2010-2012. He is currently an Associate Professor in the Institute of Automation, Chinese Academy of Sciences. His research interests include computer vision, pattern recognition, with a focus on image and video analysis, particularly face recognition, object detection, person re-identification, metric learning and video surveillance. He was awarded the Motorola Best Student Paper award and the 1st Place Best Biometrics Paper award at the International Conference on Biometrics in 2006 and 2007, respectively, for his work on face recognition. He was also awarded the best reviewer award in IJCB 2014.
\end{IEEEbiography}

\begin{IEEEbiography}[{\includegraphics[width=1in,height=1.25in,clip,keepaspectratio]{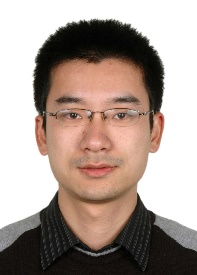}}]{Zhen Lei} received the B.S. degree in automation from the University of Science and Technology of China, in 2005, and the Ph.D. degree from the Institute of Automation, Chinese Academy of Sciences, in 2010, where he is currently an Associate Professor. He has published over 100 papers in international journals and conferences. His research interests are in computer vision, pattern recognition, image processing, and face recognition in particular. He served as an area chair of the International Joint Conference on biometrics in 2014, the IAPR/IEEE International Conference on Biometric in 2015, 2016 and 2018, and the IEEE International Conference on Automatic Face and Gesture Recognition in 2015.
\end{IEEEbiography}

\begin{IEEEbiography}[{\includegraphics[width=1in,height=1.25in,clip,keepaspectratio]{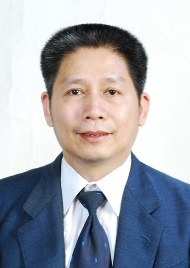}}]{Canhui Cai} received the B.S. degree from Xidian University, Xi¡¯an, China in 1982, M.S. degree from Shanghai University, Shanghai, China in 1985, and Ph.D. degree from Tianjin University, Tianjin, China in 2003, all in electronic engineering. Since 1984, he has been with the faculty of Huaqiao University, Quanzhou, China, and is currently a Professor at the College of Engineering. He was a Visiting Professor at Delft University of Technology, Delft, the Netherlands from 1991 to 1992, and a Visiting Professor at University of California at Santa Barbara, Santa Barbara, CA from 1999 to 2000. His research areas include video communications, image and video signal processing, and computer vision. He is the author or co-author of 4 books, and has published over 150 papers in journals and conference proceedings. Dr. Cai was a General Co-Chair of Intelligent Signal Processing and Communication Systems in 2007..
\end{IEEEbiography}

\begin{IEEEbiography}[{\includegraphics[width=1in,height=1.25in,clip,keepaspectratio]{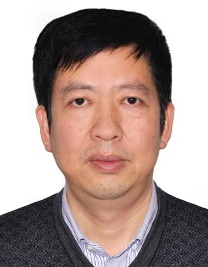}}]{Lixin Zheng} received the B.S. degree in electronic technology application and the M.S. degree in machinery manufacturing from the Huaqiao University, in 1987 and 1990, respectively. He received the the Ph.D. degree in measurement technology and instrument from Tianjin University in 2002. He is a Professor and the Dean of the College of Engineering, Huaqiao University, Quanzhou, China. His current research interests include motion control, power supply technology, computer vision and pattern recognition.
\end{IEEEbiography}

\end{document}